\documentclass[letterpaper, 10 pt, conference]{ieeeconf}  

\IEEEoverridecommandlockouts                              

\usepackage{amssymb}  
\usepackage[T1]{fontenc}
\usepackage{mathrsfs}
\usepackage{amsfonts}
\usepackage{bm}
\usepackage{amsmath}
\usepackage{microtype}
\usepackage{url}
\usepackage{booktabs}
\usepackage{tabularx}
\usepackage{multirow}

\usepackage{graphicx}
\usepackage{xcolor}
\usepackage{cite}
\usepackage[ruled,linesnumbered]{algorithm2e}

\title{\LARGE \bf
Coarse-to-Fine Detection of Multiple Seams for Robotic Welding}

\author{Pengkun Wei, Shuo Cheng, Dayou Li, Ran Song, Yipeng Zhang, and Wei Zhang${^*}$
\thanks{Pengkun Wei, Shuo Cheng, Dayou Li, Ran Song, and Wei Zhang are with the School of Control Science and Engineering, Shandong University, Jinan 250061, China}%
\thanks{Yipeng Zhang is with the Department of Electrical and Computer Engineering, University of California, Los Angeles, CA, United States of America
        {\tt\small zyp5511@g.ucla.edu}}%
\thanks{*Corresponding author: Wei Zhang (email: davidzhang@sdu.edu.cn)}%
}

\begin{document}
\begin{sloppypar}

\maketitle
\thispagestyle{empty}
\pagestyle{empty}

\begin{abstract}
Efficiently detecting target weld seams while ensuring sub-millimeter accuracy has always been an important challenge in autonomous welding, which has significant application in industrial practice. Previous works mostly focused on recognizing and localizing welding seams one by one, leading to inferior efficiency in modeling the workpiece. This paper proposes a novel framework capable of multiple weld seams extraction using both RGB images and 3D point clouds. The RGB image is used to obtain the region of interest by approximately localizing the weld seams, and the point cloud is used to achieve the fine-edge extraction of the weld seams within the region of interest using region growth. Our method is further accelerated by using a pre-trained deep learning model to ensure both efficiency and generalization ability. The performance of the proposed method has been comprehensively tested on various workpieces featuring both linear and curved weld seams and in physical experiment systems. The results showcase considerable potential for real-world industrial applications, emphasizing the method's efficiency and effectiveness. Videos of the real-world experiments can be found at \url{https://youtu.be/pq162HSP2D4}.
\end{abstract}

\section{Introduction}
According to data published by the International Federation of Robotics (IFR), welding robots account for the second largest market share of industrial robots globally. 
In recent years, although there are studies on in-welding monitoring of weld pool dynamics \cite{b3} and post-welding quality inspection \cite{c1,c2}, the majority of research still primarily focuses on pre-welding seams Recognition and localization \cite{a1,a2,a3,a5,a7,a8,zhou2021path,force1}. 

However, most of these methods are designed for a single weld seam to ensure accuracy, in other words, the sensor's field of view contains information for only one weld seam at a time. It implies that in industrial practice, the position information of all weld seams on a workpiece can only be determined one by one. In this process, multiple manual demonstrations of scanning positions are required, which leads to low efficiency.

Laser sensors are widely used in weld seam detection with the advantage of high accuracy. However, considering the limited receptive field of laser sensors, we recommend using structured light vision sensors with a larger field of view to improve detection efficiency. A larger field of view implies a larger volume of point cloud data and potentially lower accuracy of the raw data. Improving the detection efficiency while maintaining the recognition accuracy is the other key challenge. 

In order to address the aforementioned issues, in this paper we introduce a multiple weld seam edges extraction algorithm based on an RGB-D camera that can simultaneously detect all weld seams within a larger field of view while ensuring accuracy. The method consists of a coarse positioning module based on deep learning methods and a fine-edge extraction module using region growth. 
In our experiments, we found that only the point cloud surrounding the weld seams contains relevant information, while a significant amount of computational resources is wasted on other redundant parts. The deep learning module utilizes semantic segmentation results from the RGB images processed by the Fast Segment Anything Model (FastSAM) to obtain the approximate positions of the weld seam edges and crops the raw point cloud accordingly. 
The region growing module first downsamples the cropped point cloud to ensure further efficiency while preserving as much geometric information as possible. Then, the algorithm will extract all weld seams from the preprocessed point cloud at once and generate a welding path with 6-DOF. Finally, the robot arm welding system performs the execution. 
The contributions of this paper are as follows:
\begin{itemize}
    \item A redundant point cloud information removal algorithm based on FastSAM segmentation information applied to weld seam edges extraction is proposed
    \item Development of a region-growing edge extraction algorithm that can be used to extract and fit workpiece weld seams and generate weld paths that meet the requirements of sub-millimeter welding accuracy.
    \item The effectiveness of the algorithm was validated on physical experimental systems in both laboratory and industry scenarios.
\end{itemize}

\section{Related Work}
The entire welding process can be divided into several subtasks, including weld detection\cite{a1,a2,a3,a5,a8,a10}, weld tracking\cite{a4,a6,a9,b1,d1,d4,d5}, melt pool monitoring\cite{b3} and weld defect inspection\cite{c1,c2,d2}. 
The detection and location of weld seams, which serve as a pivotal and initial step in the welding process, directly affect the quality of welding. 
\subsection{Laser sensor-based Methods}
Due to the challenges involved in performing weld seam detection and localization on the entire workpiece, most methods are targeted at single weld seam. Laser sensors are widely used in the welding field relying on their high precision characteristics. 
Zhang et al. \cite{a1} proposed a grayscale image features extraction method based on laser structure light and reconstructed the curve seam model using a cubic spline smoothing technique. 
Ma et al. \cite{a2} proposed a novel self-updating template matching (SUTM) method for circular welds to extract features and an algorithm for spatial circle center fitting based on random sampling consensus (RANSAC). 
Fan et al. \cite{a3} developed a seam tracking system specifically designed for narrow seams with widths less than 0.2mm. The system performs laser scanning at multiple pre-set positions to identify the position of the narrow seam. 
For laser sensor images, Li et al. \cite{d4} proposed a welds extraction method based on the improved target detection model CenterNet, and Zhang et al.\cite{d5} presented a deep neural network-based weld feature point extraction method, which were used for seam tracking and posture adjustment. 
Although the methods based on laser sensors have the advantage of small errors, the disadvantage of narrow perception range usually restricts them to a single weld seam, which significantly reduces the efficiency.

\subsection{Other sensor-based Methods}
An edge density-based welding seam edges extraction method using an RGB-D camera was proposed in \cite{zhou2021path}. In addition to visual sensors, Ganguly and Khatib \cite{force1} also proposed a welding seam edge detection algorithm through tactile exploration with sub-millimeter accuracy. However, its prolonged exploration time and computational time are highly unfavorable for industrial assembly line production. Efficiently and accurately identifying all the weld seams to be welded has become a pressing challenge in the industrial field, demanding an urgent solution. 

Next, we will introduce a method using an RGB-D camera for extracting multiple weld seam edges. The method uses a coarse-to-fine framework to efficiently detect weld seams with comparable accuracy. Furthermore, we deployed our algorithm on different physical systems for various workpieces and achieved the desired results.


\begin{figure}[t]
	\centering
	\includegraphics[width=0.45\textwidth]{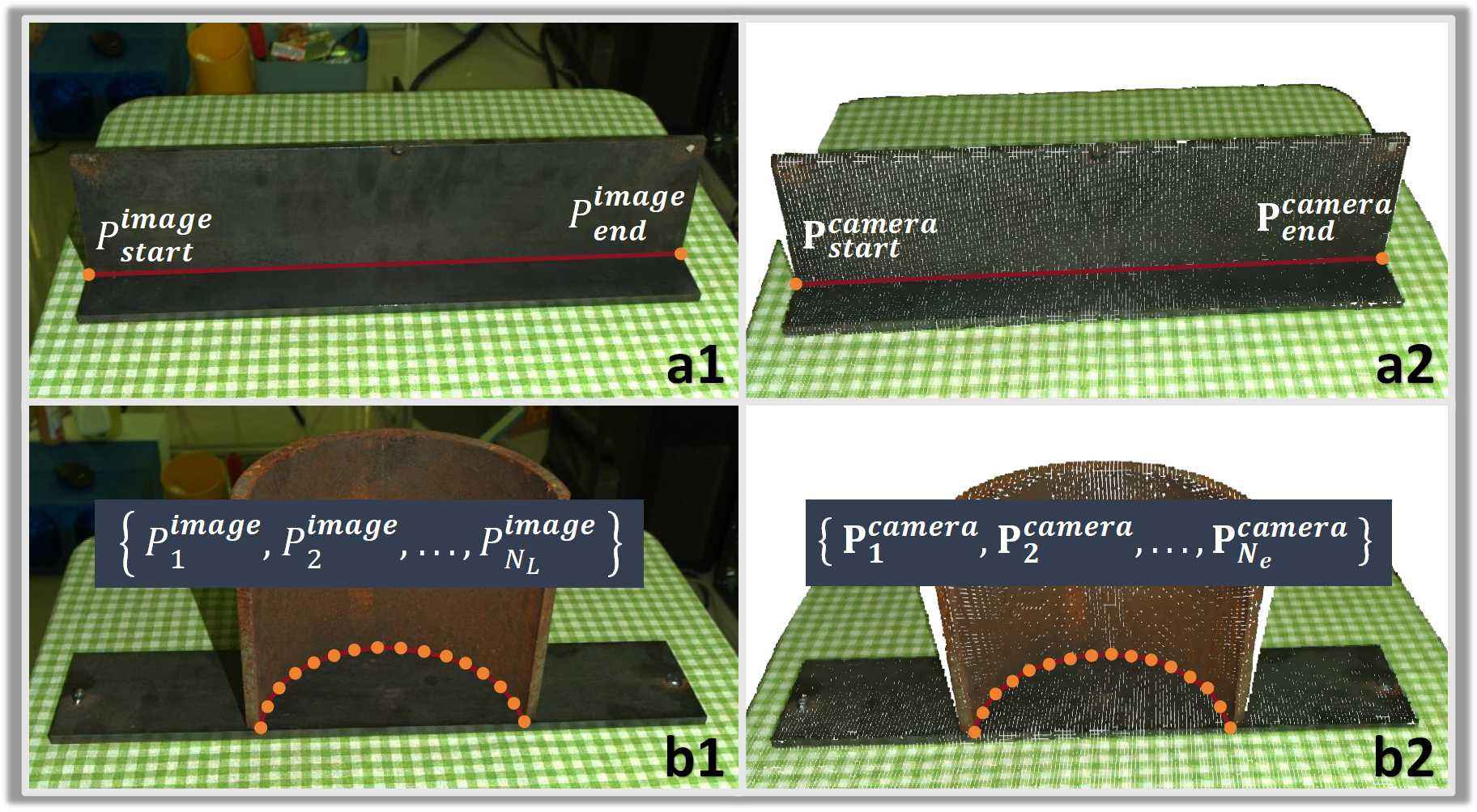} %
	\caption{(a1) An RGB image of a linear seam model; (a2) A point cloud of a linear seam model; (b1) An RGB image of a curved seam model; (b2) A point cloud of a curved seam model.}
	\label{model}
\end{figure}

\begin{figure*}[t]
	\centering
	\includegraphics[width=\textwidth]{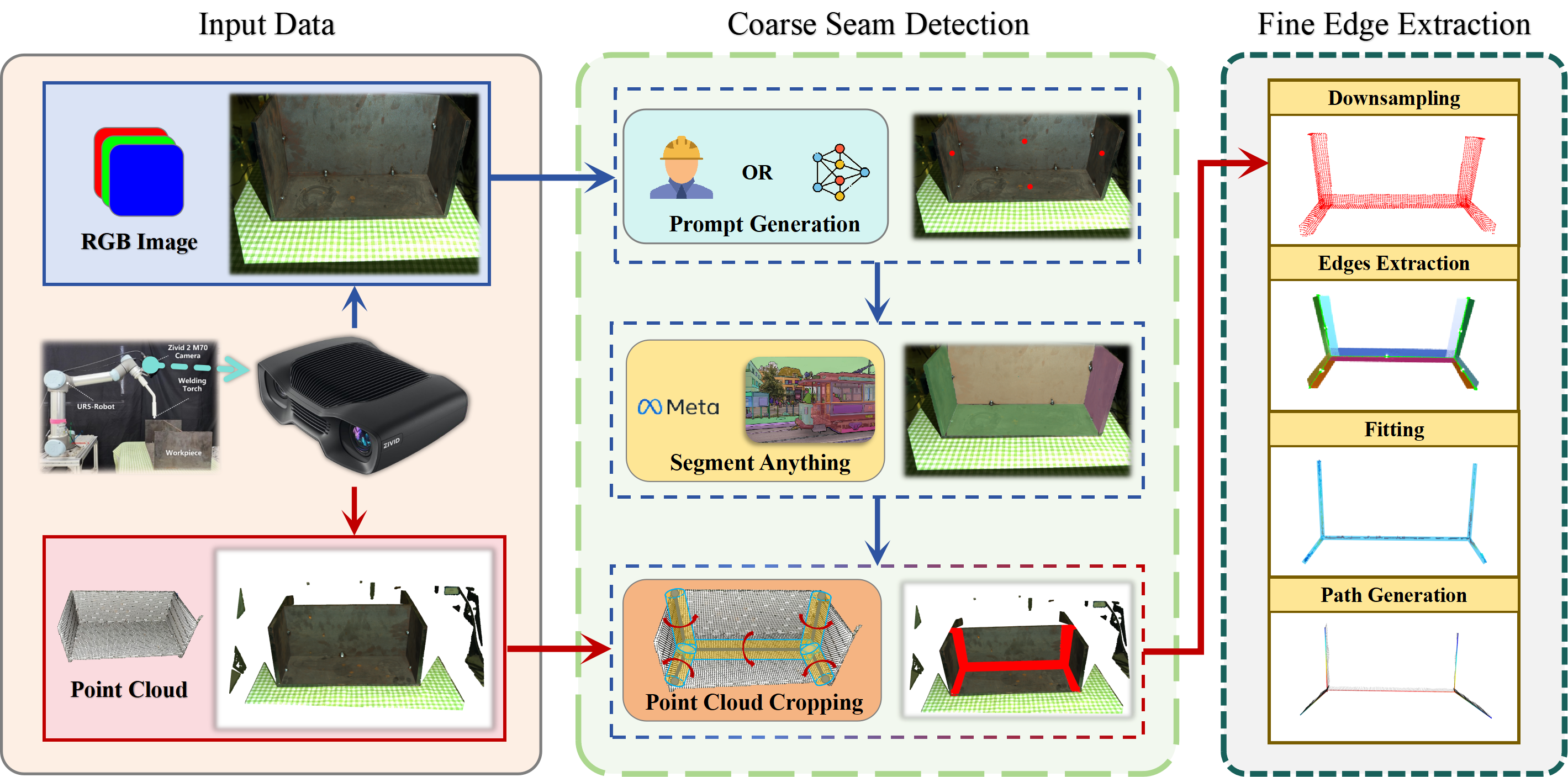} %
	\caption{Overview of Coarse-to-Fine Detection of Multiple Seams via Edge Extraction for Robotic Welding}
	\label{algorithms}
\end{figure*}

\section{Method}
\subsection{Problem Statement} 
The input of the whole system is an RGB image and raw point cloud of the workpiece acquired in the same shooting posture, while the output is a robot-executable welding path 
$\mathbb{W}\ \{\textbf{$\Psi$}_{1},\ \textbf{$\Psi$}_{2},\ \cdots,\ \textbf{$\Psi$}_{n}\}$
with 6-DOF where each $\textbf{$\Psi$}_{i}$ is represented by [ $\textbf{P}_{i},\ \textbf{R}_{i}$ ] in the world coordinate system. $\textbf{P}_{i}$ ($\textit{x}$, $\textit{y}$, $\textit{z}$) represents a point position in the world coordinate system and $\textbf{R}_{i}$ ($\textit{w}$, $\textit{p}$, $\textit{r}$) represents the end-effector pose of the robot. 



\subsection{Workpiece Edge Model} 
In the field of welding, a weld seam typically refers to the junction position between two mental plates to be welded, and the plates could be flat or curved. Types of weld seams include butt weld, lap weld, fillet weld, and T-joint weld, etc. Each type of weld seam can further be subdivided into many specific types. However, based on the geometry of the weld seams, they can generally be classified into linear and curved weld seams shown in Fig. \ref{model}. Next, we will model these two cases separately.

\textbf{Linear Weld Seams.}
Linear weld seams are the most common type of weld seam structure. We set the world coordinate system to completely coincide with the base coordinate system of the robotic arm. Typically, we use the starting point $\textbf{P}_{start}^{world}$ ($\textit{x}$, $\textit{y}$, $\textit{z}$) and the ending point $\textbf{P}_{end}^{world}$ ($\textit{x}$, $\textit{y}$, $\textit{z}$) in the world coordinate system to represent a straight-line weld seam shown in Fig. \ref{model}. We assume that the adjacent regions on both sides of the weld seam can be abstracted as C2-continuous surfaces $\mathcal{S}_1$ and $\mathcal{S}_2$, i.e., the curvature remains constant within the neighborhood except for the abrupt change in the weld seam region itself.

In RGB images, according to the mapping relationship from three-dimensional to two-dimensional space, a straight line in space is also mapped as a straight line on the two-dimensional plane. We will represent a linear weld seam in the image coordinate system using the starting pixel 
$\textit{P}_{start}^{image}$ ($\textit{x}$, $\textit{y}$) 
and the ending pixel 
$\textit{P}_{end}^{image}$ ($\textit{x}$, $\textit{y}$). 
Furthermore, we can obtain the weld seam's starting point 
$\textbf{P}_{start}^{camera}$ ($\textit{x}$, $\textit{y}$, $\textit{z}$) 
and the weld seam's ending point 
$\textbf{P}_{end}^{camera}$ ($\textit{x}$, $\textit{y}$, $\textit{z}$) 
in the camera coordinate system. Then the spatial correspondence between any point $\textbf{P}^{camera}_i$ in the camera coordinate system and its corresponding point $\textbf{P}^{world}_i$ in the world coordinate system is given by the following equation:

\begin{equation}
\textbf{P}^{world}_i = \textbf{M}^{world}_{tool} \cdot \textbf{M}^{tool}_{camera} \cdot \textbf{P}^{camera}_i
\end{equation}
where $\textbf{M}^{tool}_{camera}$ denotes the rotation and translation matrix from the camera to the tool obtained through hand-eye calibration, while $\textbf{M}^{world}_{tool}$ is the transformation matrix from the tool coordinate system to the world coordinate system, which can usually be easily obtained from the teach pendant of the robotic arm.

In the acquired point cloud data, we use a set of 
$\mathit{N}_{L}$ points $\mathbb{P}\ \{\mathcal{P}_1,\ \mathcal{P}_2,\ \cdots,\ \mathcal{P}_{\mathit{N}_{L}}\}$ 
in the neighborhood of the weld seam to fit a straight line which is represented by [ $\mathcal{P}_{start},\ \mathcal{P}_{end}$ ]. Since the point cloud data is also represented in the camera coordinate system, the relationship from the camera coordinate system to the world coordinate system for $\mathcal{P}_{start}$ and $\mathcal{P}_{end}$ also satisfies Eq. (1).

\textbf{Curved Weld Seams.}
We assume that each curved weld seam $\mathcal{E}$ is composed of $\mathit{N}_{C}$ edge segments 
\{$\mathit{e}_1,\ \mathit{e}_2,\ \cdots,\ \mathit{e}_{\mathit{N}_{C}}$\} 
that are C2-continuous, and its two side regions can be similarly abstracted into two C2 continuous surfaces. For any curve edge segment $\mathit{e}_{\mathit{i}}$,  we represent it in RGB images using a set of pixel points denoted as
$\Omega_{image}\ \{\textit{P}_{1}^{image},\ \textit{P}_{2}^{image},\ \cdots,\ \textit{P}_{N_e}^{image}$\} , corresponding to 
$\Omega_{camera}\ \{\textbf{P}_{1}^{camera},\ \textbf{P}_{2}^{camera},\ \cdots,\ \textbf{P}_{N_e}^{camera}$\} in the camera coordinate system. We also use a set of points in the point cloud
$\mathbb{P}\ \{\mathcal{P}_1,\ \mathcal{P}_2,\ \cdots,\ \mathcal{P}_{\mathit{N}_{e}}$\} within the neighborhood of the curve to represent it like linear weld seams. For any given $\textbf{P}_{i}^{camera}$ and $\mathcal{P}_i$, they also satisfy the transformation relationship in Eq. (1).


\subsection{Coarse Seam Detection Using VLM}





During point cloud processing, we found that only the point cloud around the weld seams contains rich information, while the remaining parts contribute little to weld seam detection, resulting in a waste of computational resources. Thus, we aim to filter out the irrelevant point cloud and focus on the part containing the weld seam.
As shown in Fig. \ref{algorithms}, we propose to distill weld seam priors as coarse positioning from pre-trained VLMs to obtain useful parts of the raw point cloud and then implement a region-growing edge extraction algorithm to detect weld seams.

Segment Anything Model (SAM) \cite{kirillov2023segment} is trained on web-scale datasets with powerful image segmentation capabilities and represents a paradigm shift towards more flexible and generalized segmentation models. Moreover, the SAM model introduces a new approach to image segmentation that allows interactive, point-based user input to guide the segmentation process. SAM can accept the following prompts: bounding box, point, mask, and text. FastSAM\cite{zhao2023fast} is a novel real-time CNN-based solution for Segment Anything tasks that significantly reduces computational requirements while maintaining a competitive performance similar to SAM. Considering the computational resource limitations of practical industrial deployments, we choose FastSAM as the large model for welding segmentation to distill some useful priors. The prompts are obtained by clicking the center of each workpiece surface simply. Based on these prompts, FastSAM can segment each surface of the workpiece. The intersection area between the surfaces indicates the approximate location of the weld seams, aiding us to obtain fine-grained point clouds for subsequent processing. 

Human interaction is the most accurate way to give the required prompts for FastSAM. However, for an automated welding industrial line, manual interaction will cause a dramatic loss of productivity.
Therefore, we explore an automatic prompt generation method based on keypoint detection techniques to produce suitable click prompts. 
Compared to the prompts of the bounding box, annotating key points is much more efficient (only 4 key points need to be labeled for an open square weldment). We adopt the keypoint as the prompt for FastSAM. In practice, We annotate some data for keypoint detection algorithm training. Consequently, we can obtain fine-grained point clouds from the coarse ones to get useful weld seam information for successive processes. 

\subsection{Fine Edge Extraction Based on Region-Growing}


We propose an algorithm based on region growing to detect the weld seam edges in point cloud data. The algorithm can take either the raw point cloud data or the cropped point cloud data as input and output the position information of the weld seams. The efficiency of the execution of the algorithm varies depending on the input data, which we will explain in detail in the experiments section. The algorithm consists of the following main steps as shown in Alg. \ref{algorithm}:
\begin{itemize}
    \item the raw point cloud needs to be preprocessed, including pass-through filtering and downsampling, and building a KD-tree from the obtained point cloud to improve the efficiency of range search and nearest neighbor search. The cropped point cloud can be processed directly from downsampling.
    \item utilizing the region-growing algorithm to segment the workpiece and its individual surfaces. 
    \item obtaining the point cloud near the edges and using the least squares method to fit it. Simultaneously, calculating the welding pose to get the welding path.
\end{itemize}

\textbf{Preprocessing of The Raw Point Cloud.} The raw point cloud obtained from the camera usually contains a significant amount of noise. To extract the region of interest and remove irrelevant points, we apply a pass-through filter to the point cloud based on prior knowledge such as the optimal working range of the camera and the known height of the workbench. We voxel-downsample the point cloud after filtering, which divides the space into small rasters of voxels, replacing all points in the raster with the center of gravity of the point cloud within each raster. For any $\mathcal{P}_{i}$(\textit{x}, \textit{y}, \textit{z}) in the point cloud we can use the following formula to calculate the raster index (\textit{$h_x$}, \textit{$h_y$}, \textit{$h_z$}) at which the point is located.

\begin{equation}
    h_k=\left\lfloor\left(k-k_{\min }\right) / r\right\rfloor,\ k = x, y, z
\end{equation}
where \textit{r} is the side length of the raster and ($x_{\min }, y_{\min }, z_{\min }$) represents the minimum coordinates of the point cloud on each axis.

This step helps to reduce the density of the point cloud while preserving the overall shape and structural information of the workpiece. 
To further improve efficiency, we utilize the KD-tree data structure for point cloud processing. KD-tree is a data structure used in computer science to organize points in a space with k dimensions. It is commonly applied in point cloud range search and nearest-neighbor search, enabling a reduction in time complexity to $O{(n}{\log_2n}{)}$.

\textbf{Region-Growing Algorithm for Edge Extraction.} We demonstrated the entire process of region-growing segmentation in Alg. 1. The curvature of all points is calculated according to the following formula and the points are sorted based on the curvature value, then the point with the smallest curvature value is selected as the initial seed point $\textbf{P}_{inatial}$ to be added to the seed set $\mathcal{S}_{seed}$. According to the edge model established earlier, the smoother the region in the point cloud, the less likely it is to be a weld region, and growing from the smoothest region could improve efficiency.

\begin{equation}
\begin{aligned}
    \mathrm{M}&=\frac{1}{\mathrm{k}} \sum_{\mathrm{i}=1}^{\mathrm{k}}\left(\mathcal{P}_i-\mathcal{P}_0\right)\left(\mathcal{P}_i-\mathcal{P}_0\right)^{\mathrm{T}} \\
    \delta&=\frac{\lambda_0}{\lambda_0+\lambda_1+\lambda_2} \\
\end{aligned}
\end{equation}
where $\mathrm{M}$ donates the covariance matrix of the neighborhood formed by the $\mathrm{k}$ nearest points to $\mathrm{p}_{\mathrm{i}}$, $\mathrm{p}_0$ represents the barycenter of the neighborhood, ${\lambda_0}, {\lambda_1}, {\lambda_2}$ represents the three eigenvalues of the covariance matrix M, and ${\lambda_0}<{\lambda_1}<{\lambda_2}$, then a smaller value of $\delta$ indicates a flatter neighborhood, whereas a more rugged neighborhood.


\begin{algorithm}[t]
\SetAlgoLined 
\caption{Region-Growing Segmentation}\label{algorithm}

\KwIn{input parameters A, B, C}
\KwOut{output result}

\If{$\mathcal{S}_{seed}$ is not empty}{

    \For{$\textbf{P}_{inatial}$ in $\mathcal{S}_{seed}$}{
    
        $\mathcal{S}_{neighborhood}$ = $\textbf{NeighborSearch}$ ( $\textbf{P}_{inatial}$ )
        
        \For{$\textbf{P}_{neighbor}$ in $\mathcal{S}_{neighborhood}$}{
        
            Angle = $\textbf{VectorsAngle}$ ( $\textbf{P}_{neighbor}$ )
            
            Curvature = $\textbf{CalCurvature}$ ( $\textbf{P}_{neighbor}$ )

            \eIf{Angle > Threshold1}{

            $\mathcal{S}_{edges}$ $\leftarrow$ $\textbf{P}_{neighbor}$
            
            }{
            
            $\mathcal{S}_{surfaces}$ $\leftarrow$ $\textbf{P}_{neighbor}$
            }

            \If{Curvature < Threshold2}{
            $\mathcal{S}_{seed}$ $\leftarrow$ $\textbf{P}_{neighbor}$

            $\mathcal{S}_{seed.}$$\textbf{Delete}$($\textbf{P}_{inatial}$)
            }
        }
    }
}

\end{algorithm}

\begin{figure*}[t]
	\centering
	\includegraphics[width=\textwidth]{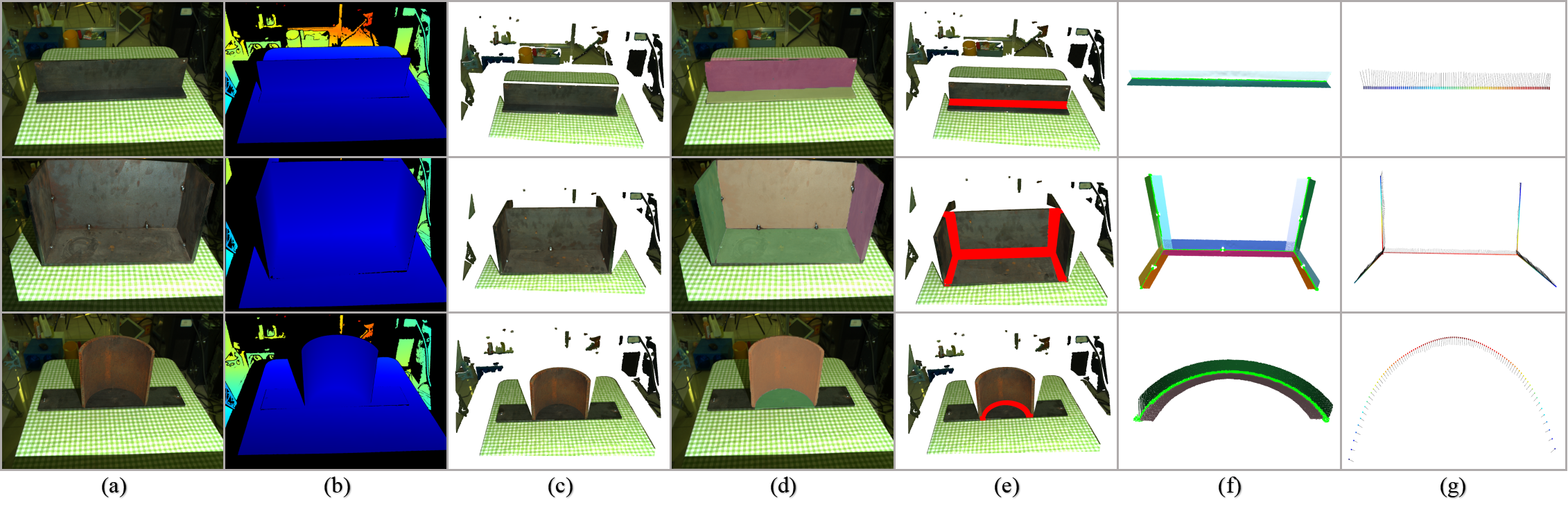} %
	\caption{Results in different processes for three workpieces. (a) RGB Images; (b) Depth Maps; (c) Raw Point Cloud; (d) Segmentation Results; (e) The red area represents the point cloud cropped according to the segmentation information; (f) The light green area represents the extracted weld edges point cloud; (g) Robot-executable Welding Path with 6-DOF.}
	\label{results}
\end{figure*}
\begin{figure}[t]
	\centering
	\includegraphics[width=0.45\textwidth]{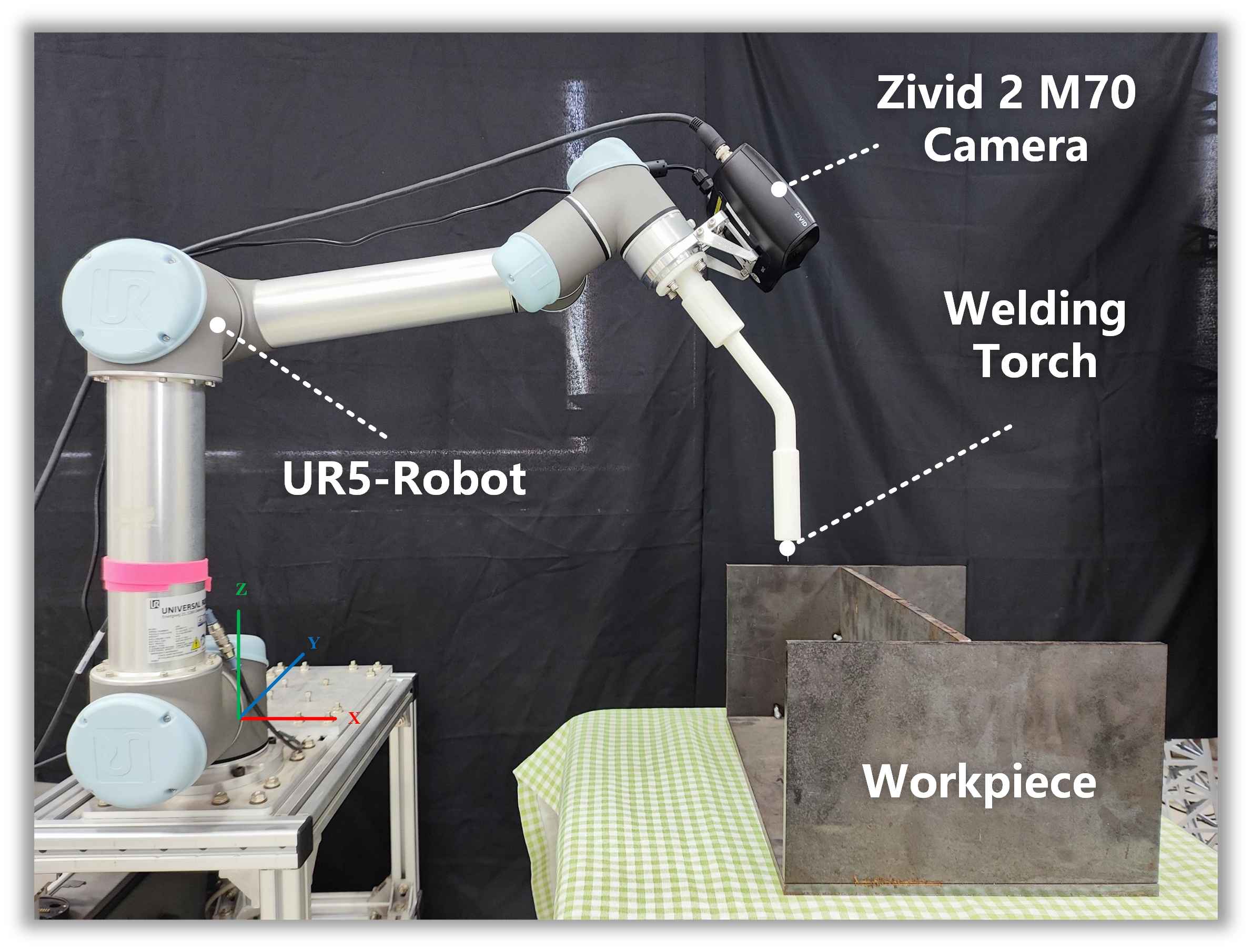} %
	\caption{A robotic welding system consists of a UR5 robotic arm, an RGB-D camera, a 3D-printed welding torch, and a workpiece to be welded.}
	\label{system2}
\end{figure}

\begin{figure}[t]
	\centering
	\includegraphics[width=0.45\textwidth]{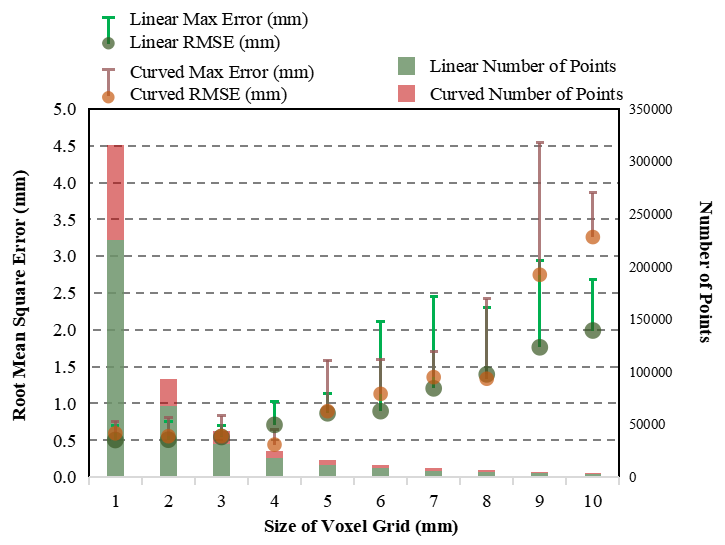} %
	\caption{The changes in the quantity of downsampled point clouds and the corresponding variations in the maximum error and root mean square error (RMSE) resulting from selecting different grid sizes.}
	\label{downsampling}
\end{figure}
Then we can divide the point cloud into two parts: the part containing edges $\mathbb{P}_{edges}$ and the part not containing edges $\mathbb{P}_{surfaces}$. The next step is to extract the point cloud that accurately represents the positions of the edges from $\mathbb{P}_{edges}$. Based on the previously established edge model, the welding seam edges not only need to have high curvature but also need to represent the intersection between two surfaces. Therefore, we perform a nearest neighbor search on the edge point cloud to find points that satisfy both conditions, which correspond to the precise point cloud representing the location of the welding edge.

\textbf{Welding Path Generation.} Next, we need to fit the precise point cloud of the weld seam edge in order to generate the welding path. We first use the weld edge point cloud to fit the ${Plane}$ [\textit{A, B, C, D}] it lies on, then project the point cloud onto this plane, and finally construct a rotation matrix to convert any projected point cloud 
$\textbf{P}^{camera}$ ($\textit{x}$, $\textit{y}$, $\textit{z}$) 
in the camera coordinate system to 
$\textbf{P}^{plane}$ ($\textit{x}$, $\textit{y}$, $\textit{z}$) 
in the planar coordinate system, which has a z-axis perpendicular to the plane itself. So the $\textit{z}$ after the conversion is a fixed value, and we only need to fit the 
($\textit{x}$, $\textit{y}$). 
The weld path contains the weld trajectories and the corresponding welding torch poses. The welding trajectories can be obtained by uniformly sampling points from the fitted edge. The selection of welding torch poses follows two principles: one is not to make the welding torch collide with the workpiece, and the other is to make the welding torch as perpendicular as possible to the tangent line of the welded place in order to ensure the quality of the weld.
We take points within the field of the weld point cloud obtained in the previous step and compute their mean vector direction as the welding pose.



\section{Experiments}
In order to validate the proposed method, we conducted several experiments on a robotics welding system based on a UR5 robotic arm shown in Fig.~\ref{system2}. Next, we evaluated the system by performing a series of experiments on two linear-weld-seam workpieces and one curved-weld-seam workpiece. It is worth mentioning that we have also attempted to deploy our method in real industrial welding scenario and achieved promising results.


\subsection{Effect of the Downsampling Parameter}
As the sensor field of view increases, the number of point clouds to be processed also increases dramatically. 
When dealing with point cloud data, it is common to use downsampling as a means to reduce the density of point clouds. 
Downsampling can effectively reduce the number of point clouds, thereby reducing the consumption of computational resources and speeding up subsequent processing, while still preserving as much geometric information as possible. 

In the proposed method, we employed the voxel grid downsampling method, which works by dividing the whole point cloud into regular cubic grid cells and selecting a representative point for each grid cell to represent all points within that cell.
We can adjust the edge length of the grid, and a larger grid size results in sparser point clouds and fewer points after downsampling. However, it also implies lower data precision.

\begin{table*}[t]\normalsize
\centering
\caption{Comparison of different methods in efficiency and accuracy}\label{tab_res}
\begin{tabular}{m{2.5cm}<{\centering}m{1.5cm}<{\centering}m{1.3cm}<{\centering}m{2.6cm}<{\centering}m{1.5cm}<{\centering}m{1.5cm}<{\centering}m{2.2cm}<{\centering}m{1.3cm}<{\centering}}
\toprule[1pt]
    System & Workpiece Type & Number of Welds & Method & Exploration Time (s) & Computation Time (s) & Average Time Per Weld (s) & RMSE (mm)\\  
\cmidrule(lr){1-8}
    \addlinespace[0.1cm]
    Zhou et al. \cite{zhou2021path} & Linear & 1 & Point Cloud & 10 & 3.82 & 23.16 & 1.09 \\
    \addlinespace[0.2cm]
    \multirow{2}*{Peng et al. \cite{a8}} & Linear & 1 & Point Cloud & 10 & 14.09 & 24.09 & -- \\
    & Curved & 1 & Point Cloud & 10 & 14.08 & 24.08 & -- \\
    \addlinespace[0.2cm]
    Yang et al. \cite{a10} & Both & 1 & Point Cloud & 10 & 4.72 & 14.72 & 1.18 \\
    \addlinespace[0.2cm]
    \multirow{2}*{Ganguly et al. \cite{force1}} & Linear & 1 & Tactile & 20 & 10.8 & 30.8 & 0.55 \\
    & Curved & 1 & Tactile & 20 & 35.3 & 55.3 & 0.26 \\
    \addlinespace[0.2cm]
    \multirow{3}*{Ours} & Linear & 2 & RGB+Point Cloud & 20 & 2.11 & 11.06 & 0.37 \\
    & Linear & 10 & RGB+Point Cloud & 20 & 10.54 & 3.05 & 0.54 \\
    & Curved & 1 & RGB+Point Cloud & 10 & 1.382 & 11.38 & 0.56 \\
    
\bottomrule[1pt]
\end{tabular}
\end{table*}

We conducted experiments to illustrate how to choose the grid size to balance accuracy and efficiency. As shown in Fig. \ref{downsampling}, we performed experiments on a weldment containing five linear weld seams and a curved weldment, and sequentially tested the variation of point cloud quantity and error as the voxel grid size changed from 1mm to 10mm.
The root mean square error (RMSE) was used to represent the gap between the path points generated by our method and the ground truth points which were obtained by manually annotating high-precision point clouds. 
We uniformly selected 3 points on each linear weld seam and 10 points on the curved weld seam. The average errors were calculated using the following formula:

\begin{equation}
RMSE=\sqrt{\frac1n\sum_{i=1}^n\sum_k\left(\mathcal{P}_m-\mathcal{P}_g\right)^2}
\end{equation}
where $\mathcal{P}_m$ represents the generated welding path point by our method, $\mathcal{P}_g$ is the ground truth of the welding path point, \textit{k} means the x, y, and z dimensions of each path point, and \textit{n} denotes the number of path points.
According to welding requirements, the maximum error of the welding path should not exceed 1 mm. From the experimental results, it is observed that setting the edge length to 3 mm can ensure sub-millimeter accuracy while maximizing efficiency.

\subsection{Quantitative Comparison}

\begin{table}[t]\normalsize
\centering
\caption{Comparison results with and without cropping}\label{tab_deep}
\begin{tabular}{m{3cm}<{\centering}m{2.5cm}<{\centering}m{1.5cm}}
\toprule[1pt]
    Method & Number of Points & Time (s) \\  
\cmidrule(lr){1-3}
    Point Cloud & 81730 & 33.51 \\
    RGB+Point Cloud & 10356 & 10.54 \\
\bottomrule[1pt]
\end{tabular}
\end{table}

Zhou Peng et al.\cite{zhou2021path} proposed a novel point cloud intensity-based method to extract edge points and generate weld paths. Peng Rui et al.\cite{a8} proposed a welding groove detection algorithm and conducted experiments to validate the accuracy and efficiency of the algorithm. Yang et al.\cite{a10} designed a fringe projection system to measure the appearance of welding workpieces and proposed a 3D seam extraction algorithm based on point cloud segmentation. The previous methods are all based on point cloud, Ganguly et al.\cite{force1} proposed a method for robotic welding through tactile exploration. Next, We will compare our method with these baselines on two types of workpieces.

We prefer to take the time of the entire process into account as opposed to comparing the algorithm runtime alone.
Thus, we divide the total time cost into exploration time and computation time. Exploration time refers to the time spent on acquiring the input data to be processed by the sensors, while computation time refers to the time taken for the algorithm to perform. Since most existing methods have test data for a single weld seam, our proposed method is capable of extracting multiple weld seams simultaneously. In the end, we will compare the average time for each weld seam. To ensure fairness, the exploration time for all methods, except the tactile-exploration-based method, is set to ten seconds per capture. The exploration time includes the time for teaching the capture positions and the time taken for actual capturing.

According to the experimental data shown in \ref{tab_res}, we achieved the highest efficiency in terms of average time spent on the two types of workpieces. It is worth noting that our average time decreases as the number of weld seams captured in each shot increases.

To validate the effectiveness of our proposed FastSAM-based coarse localization cropping method, we design comparative experiments to compare the number of point clouds that the algorithm needs to process and the time required with and without the inclusion of the cropping module. Taking the workpiece containing ten linear weld seams as an example, the experimental results demonstrate that adding the cropping module can remove over 80\% of redundant point clouds and save approximately 68.5\% of computational time.

\subsection{Application in Industry Scenario}

\begin{figure}[t]
	\centering
	\includegraphics[width=0.48\textwidth]{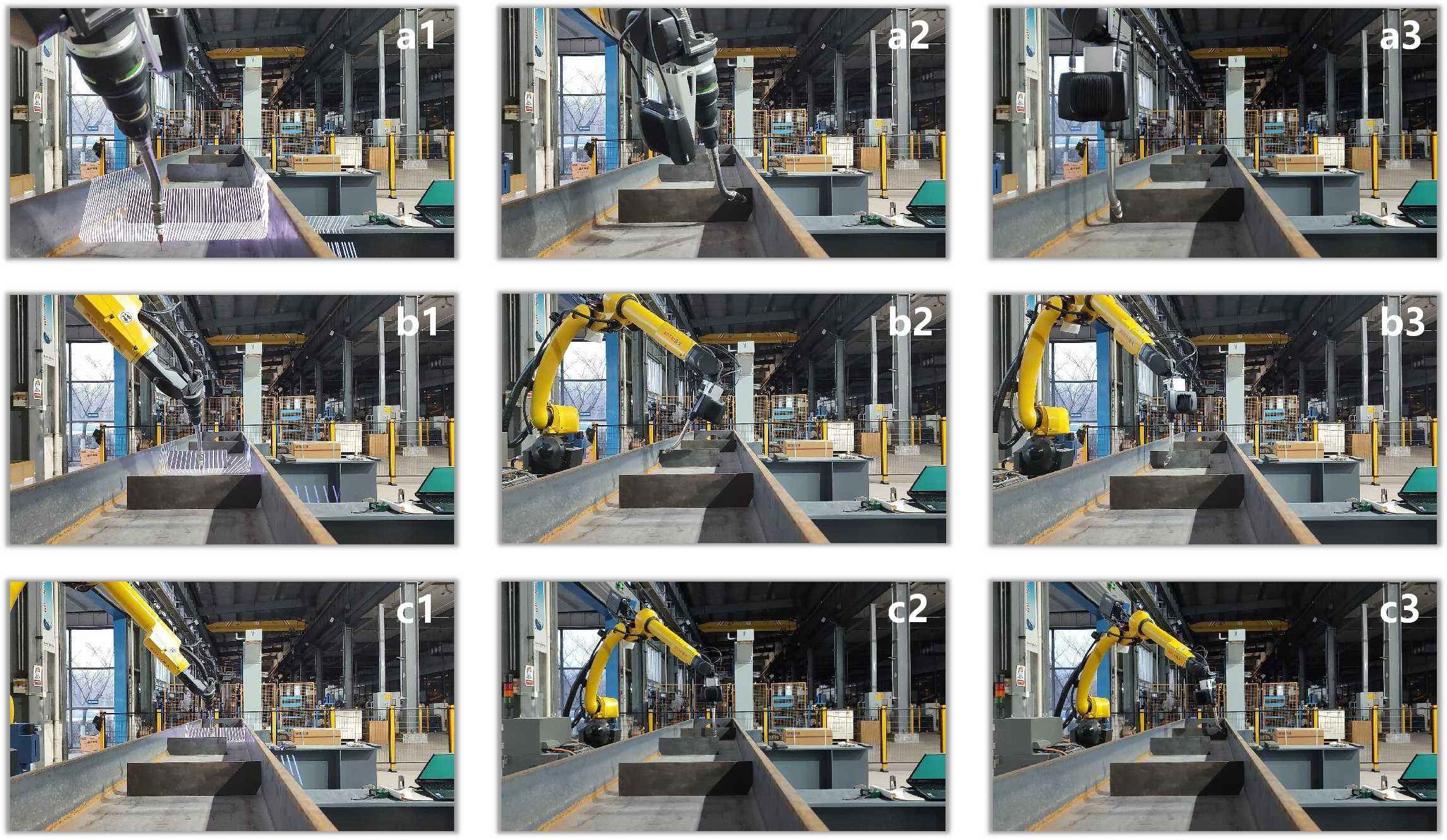} %
	\caption{By collaborating with the ground track, the robotic arm can accurately identify and weld each set of weld seams on the workpiece. a1-a3, b1-b3, and c1-c3 represent different types of connections between the rib plate and the main body of the workpiece, respectively.}
	\label{video}
\end{figure}

To further validate the effectiveness of the proposed algorithm in real industrial scenarios, we conducted experiments on a welding system based on the Fanuc M-10iD/8L robot arm equipped with a ground track. The workpieces to be welded are conventional steel structures, which are widely applied in various fields such as bridges, buildings, and ships, offering broad application prospects. The task is to weld the ribs to the main body of the workpiece, where the ribs play a role in enhancing structural strength and providing support. Depending on the requirements, the ribs can be connected to the main body at either one end or both ends. As shown in Fig. \ref{video}, We included these cases in our experiments. Video demonstration of the real-world tests can be found at \url{https://youtu.be/pq162HSP2D4}.

\section{Conclusions}
This paper presents a multi-weld-seam detection method applied in the field of robotic welding. For the RGB images and point cloud data acquired by sensors, a coarse-to-fine weld seam edge extraction algorithm is designed and tested in both laboratory and factory scenarios. We quantitatively compare the proposed method with existing methods in terms of accuracy and efficiency on different linear and curved workpieces. The experimental results demonstrate that our method can improve the efficiency of detecting weld seams while ensuring comparable accuracy. Regarding the voxel grid downsampling method, we designed experiments to demonstrate how to select an appropriate grid size for the voxel grid.

In the future, we will introduce the point cloud reconstruction method to reconstruct the more complex weldments such as those containing both straight and curved weld seams, and we will try to extract the weld seams in the reconstructed point cloud.

\addtolength{\textheight}{-12cm}   











\bibliographystyle{unsrt}
\bibliography{ref}
\end{sloppypar}
\end{document}